\documentclass[a4paper,orivec]{llncs}

\newif\ifdraft\drafttrue
\newif\ifdotikz\dotikzfalse

\draftfalse

\usepackage{a4,a4wide}
\usepackage{times}
\usepackage{helvet}
\usepackage{courier}
\usepackage{xspace}
\usepackage{proof}
\usepackage{ccg}
\usepackage{url}
\usepackage{amsmath}
\usepackage{amssymb}
\usepackage{color}
\usepackage{graphicx}
\usepackage{paralist}
\usepackage{cite}

\graphicspath{{./}}
\ifdotikz
\usepackage{tikz}
\pgfrealjobname{aspccg}
\usetikzlibrary{shapes,arrows,backgrounds,%
matrix,patterns,arrows,decorations.pathmorphing,decorations.pathreplacing,%
positioning,fit,calc,decorations.text,shadows%
}
\else
\long\def\beginpgfgraphicnamed#1#2\endpgfgraphicnamed{\includegraphics{#1}}
\fi

\ifdraft
\usepackage{comments}
\else
\newcommand{\comment}[1]{}
\fi

\newenvironment{myitemize}{\begin{list}{$\bullet$}{%
\setlength{\topmargin}{0pt}
\setlength{\leftmargin}{0pt}
\setlength{\itemindent}{10pt}}
}{\end{list}}

\def\beq{\begin{equation}}
\def\eeq#1{\label{#1}\end{equation}}

\def\ba{\begin{array}}
\def\ea{\end{array}}

\def\ar{\leftarrow}

\def\bi{\begin{itemize}}
\def\ei{\end{itemize}}

\def\ccgasp{{\tt ccg.asp}\xspace}
\def\ccgtoidpdrawasp{{\tt ccg2idpdraw.asp}\xspace}

\newcommand{\mi}[1]{\ensuremath{\mathit{#1}}}

\def\NP{\ensuremath{\mi{NP}}}

\def\id{\ensuremath{\mi{id}}}
\def\posCat{\ensuremath{\mi{posCat}}}
\def\posAdjacent{\ensuremath{\mi{posAdjacent}}}
\def\posLastAffected{\ensuremath{\mi{posLastAffected}}}
\def\posAffected{\ensuremath{\mi{posAffected}}}
\def\ban{\ensuremath{\mi{ban}}}
\def\occurs{\ensuremath{\mi{occurs}}}
\def\mytime{\ensuremath{\mi{time}}}
\def\rfunc{\ensuremath{\mi{rfunc}}}
\def\lfunc{\ensuremath{\mi{lfunc}}}
\def\unary{\ensuremath{\mi{unary}}}
\def\binary{\ensuremath{\mi{binary}}}

\def\maxsteps{\ensuremath{\mi{maxsteps}}}
\def\ruleFwdAppl{\ensuremath{\mi{ruleFwdAppl}}}
\def\ruleBwdRaise{\ensuremath{\mi{ruleBwdRaise}}}
\def\TLast{\ensuremath{\mi{TLast}}}

\def\gringo{{\sc gringo}\xspace}
\def\clasp{{\sc clasp}\xspace}

\def\candc{C\&C\xspace}
\def\openccg{{\sc openCCG}\xspace}
\def\lkb{LKB\xspace}
\def\tccg{TCCG\xspace}
\def\idpdraw{IDPDraw\xspace}
\def\ccgbank{CCGbank\xspace}
\def\aspccg{{\sc AspCcgTk}\xspace}
\def\boxer{Boxer\xspace}

\newcommand{\gen}{\textsc{generate}}
\newcommand{\define}{\textsc{define}}
\newcommand{\test}{\textsc{test}}

\title{Parsing Combinatory Categorial Grammar with Answer Set Programming:
Preliminary Report}

\author{Yuliya Lierler\inst{1} \and Peter Sch\"{u}ller\inst{2}}

\institute{%
Department of Computer Science, University of Kentucky\\
\email{yulia@cs.uky.edu}%
\and
Institut f\"ur Informationssysteme, Technische Universit\"at Wien\\
\email{ps@kr.tuwien.ac.at}%
}

\begin{document}

\maketitle

\newcommand{\nop}[1]{}
\begin{abstract}
Combinatory categorial grammar (CCG)
is a grammar formalism used for natural language parsing.
CCG assigns structured lexical categories to words
and uses a small set of combinatory rules
to combine these categories to parse a sentence.
In this work we propose and implement a new approach to
CCG parsing that relies on a prominent knowledge representation formalism,
answer set programming (ASP)
--- a declarative programming paradigm.
We formulate the task of CCG parsing as a planning problem
and use an ASP computational tool to compute solutions that correspond to valid parses.
Compared to other approaches, there is no need to implement a specific
parsing algorithm using such a declarative method.
Our approach aims at producing all semantically distinct parse trees
for a given sentence. %
From this goal,
normalization and efficiency issues arise, and
we deal with them by combining and extending existing strategies.
We have implemented a CCG parsing tool kit --- \aspccg --- that
uses ASP as its main computational means.
The \candc supertagger can be used as a preprocessor within \aspccg,
which allows us
to achieve wide-coverage natural language parsing.
\end{abstract}
{
\section{Introduction}

The task of parsing,
i.e., recovering the internal structure of sentences,
is an important task in natural language processing.
Combinatory categorial grammar (CCG)
is a popular grammar formalism used for this task.
It assigns basic and complex lexical categories to words in a sentence
and uses a set of combinatory rules
to combine these categories to parse the sentence.
In this work we propose and implement a new approach to
CCG parsing that relies on a prominent knowledge representation formalism,
answer set programming (ASP)
--- a declarative programming paradigm.
Our aim is to create a wide-coverage\footnote{The goal of
  wide-coverage parsing is to parse sentences that are not within a controlled fragment of natural language,
  e.g., sentences from newspaper articles.} parser
which returns all semantically distinct parse trees for a given
sentence.}

One major challenge of natural language processing is ambiguity of natural language.
For instance, many sentences have more than one plausible internal
structure, which often provide different semantics to the same sentence.
Consider
a sentence

\begin{center}
{\sl John saw the astronomer with the telescope.}
\end{center}
It can denote
that John used a telescope to see the astronomer,
or that John saw an astronomer who had a telescope.
 It is not obvious which meaning is the correct one without additional context.
Natural language ambiguity inspires our goal to return
\emph{all semantically distinct} parse trees
for a given sentence.

CCG-based systems
\openccg~%
\cite{2003_openccg}
and
\tccg~%
\cite{2003_l_documentation_a_ccg_implementation_for_the_lkb,2004_l_type_inheritance_combinatory_categorial_grammar}
(implemented in the \lkb toolkit)
 can provide multiple parse trees for a given sentence.
Both  use chart parsing algorithms
with CCG extensions such as modalities or hierarchies of categories.
While \openccg is primarily geared towards generating sentences from logical forms,
\tccg targets parsing.
However,
both implementations
require lexicons\footnote{A CCG lexicon is a mapping from each word that can occur in the input
to one or more CCG categories.}
with specialized categories.
Generally, crafting a CCG lexicon is a time--consuming task.
An alternative method to using a (hand-crafted) lexicon has been
developed and implemented in a wide-coverage CCG parser --- \candc~\cite{cla03,2004_l_parsing_the_wsj_using_ccg_and_log_linear_models}.
\candc  relies on machine learning techniques
for tagging an input sentence with CCG categories
as well as for creating parse trees with a chart algorithm.%
 As training data, \candc uses \ccgbank ---
a corpus of CCG derivations and dependency structures~\cite{2007_l_ccgbank}.
The parsing algorithm of \candc returns a \emph{single} most probable parse tree for a given sentence.

In the approach that we describe in this paper we formulate the task
of CCG parsing as a planning problem.
Then we solve it using  answer set programming~\cite{mar99,nie99}.
ASP is a declarative
programming formalism based on the answer set semantics of logic
programs~\cite{gel88}.
The idea of ASP is to represent a given computational problem by a
program whose answer sets correspond to solutions, and then use an
answer set solver
to generate answer sets for this program.
Utilizing ASP for CCG parsing allows us to control
the parsing process
with declarative descriptions of constraints on combinatory rule applications and parse trees.
  Moreover, there is no need to implement a specific parsing algorithm,
  as an answer set solver is used as a computational vehicle of the method.
Similarly to \openccg and \tccg, in our ASP approach to CCG parsing we
formulate a problem in such a way that  multiple parse trees are computed.

An important issue inherent to CCG parsing are spurious parse trees:
a given sentence may have many distinct parse trees which yield the same semantics.
Various methods for eliminating such spurious parse trees have been proposed~%
\cite{cla03,wit87,1996_l_efficient_normal_form_parsing_for_combinatory_categorial_grammar}.
We adopt some of these syntactic methods in this work.

We implemented our approach in an \aspccg toolkit.
The toolkit equips a user with two possibilities
for assigning plausible categories to words in a sentence:
it can either use a given (hand-crafted) CCG lexicon
  or it can take advantage of the \candc
  supertagger~\cite{2004_l_parsing_the_wsj_using_ccg_and_log_linear_models} for this task.
  The second possibility provides us with wide-coverage CCG parsing capabilities comparable to \candc.
  The \aspccg toolkit computes best-effort parses in cases where no
  full parse can be achieved with CCG,
  resulting in parse trees for as many phrases of a sentence as possible.
  This behavior is more robust than completely failing in producing a parse tree.
  It is also useful for development, debugging, and experimenting with rule sets and normalizations.
  In addition to producing parse trees,
  \aspccg contains a module for visualizing CCG derivations.

A number of theoretical characterizations of CCG parsing exists.
They differ in their use of specialized categories,
their sets of combinatory rules, or specific
conditions on applicability of rules.
An ASP approach to CCG parsing implemented in \aspccg
can be seen as a basis of a generic tool
for encoding different CCG category and rule sets  in a declarative
and straightforward manner. Such a tool provides a test-bed
 for experimenting with different theoretical CCG frameworks
 without the need to craft specific parsing
 algorithms.

The structure of this paper is as follows:
we start by reviewing
planning, ASP, and CCG. %
We describe our new approach to CCG parsing by formulating this task
as a planning problem
in Section~\ref{sec:ccgviaplanning}.
The implementation and framework for realizing
this approach using ASP technology is the topic of
Section~\ref{sec:implframework}.
We conclude with a discussion of future work directions and challenges.

\section{Preliminaries}\label{sec:prelims}

\subsection{Planning}\label{sec:planning}

Automated planning~\cite{cim08} is a widely studied area in Artificial
Intelligence.
In {\sl planning}, given knowledge about
\begin{enumerate}
\item[(a)] available actions, their executability, and effects,
\item[(b)] an initial state, and
\item[(c)] a goal state,
\end{enumerate}
the task is to find a sequence of
actions that leads from the initial state to the goal state.
A number of special purpose   planners
have been developed in this sub-area of Artificial
Intelligence. Answer set programming %
provides a viable alternative to special-purpose planning
tools~\cite{nie99,lif02,2004_a_logic_programming_approach_to_knowledge_state_planning_semantics_and_complexity}.

\subsection{Answer Set Programming (for Planning)}\label{sec:asp}
Answer set programming (ASP)~\cite{mar99,nie99} is a declarative
programming formalism based on the answer set semantics of logic
programs~\cite{gel88,gel91b}.
The idea of ASP is to represent a given computational problem by a
program whose answer sets correspond to solutions, and then use an
answer set solver
to generate answer sets for this program.
In this work we use the
{\clasp}\footnote{{\tt http://potassco.sourceforge.net/}.} %
system
 with its front-end (grounder)
{\gringo}~\cite{gringomanual},
which is currently one of the most widely
 used answer set solvers.

A common methodology to solve a problem in ASP is to design
 {\gen}, {\define}, and {\test}~\cite{lif02} parts of a program.
The {\gen} part defines a large collection of answer sets that could be
seen as potential
solutions.
The {\test} part consists of rules that eliminate the
answer sets that do not correspond to solutions.
The {\define} section expresses additional concepts
and connects the {\gen} and {\test} parts.

A typical logic programming rule has a form of a Prolog rule. For instance,
program
$$
\ba{l}
p.\\
q\ar p,\ \mi{not}\ r.
\ea
$$
is composed of such rules.
This program has one answer set $\{p,q\}$.
In addition to Prolog rules,
{\gringo}  also accepts rules of other kinds --- ``choice rules''
and ``constraints''. For example, rule
$$\{p,q,r\}.$$
is a choice rule. Answer sets of this one-rule program  are
arbitrary subsets of the atoms $p,\ q,\ r$. Choice rules are typically
the main members of the {\gen} part of the program.
Constraints often form
the {\test} section of a program. Syntactically, a constraint is the
rule with an empty head. It encodes the conditions on the answer sets
that have to be met. For instance,  the
constraint
$$
\ar p,\ \mi{not}\ q.
$$
eliminates the answer sets of a program that include $p$ and do not
include $q$.

System {\gringo} allows the user to specify large
programs in a compact way, using rules with schematic variables and
other abbreviations.
 A detailed description of its input language  can be found
in the online manual~\cite{gringomanual}.
 Grounder {\gringo} takes a program ``with abbreviations'' as an
input and produces its propositional counterpart that is then
processed by {\clasp}. %
  Unlike Prolog systems, the inference mechanism of {\clasp} is related to that of Propositional
Satisfiability (SAT) solvers~\cite{geb07}.

The {\gen}-{\define}-{\test}
methodology is suitable for modeling planning problems. %
To illustrate how ASP programs can be used to solve
							such problems,
we present a simplified part of the encoding of a classic toy planning
domain {\sl blocks world} given in~\cite{lif02}. In this domain,
blocks are moved by a robot. There are a number of
 restrictions  including the fact that
 a block cannot be moved unless it is clear.

Lifschitz~\cite{lif02} models the blocks world domain by means of five
predicates: {\it time/1},  {\it block/1}, {\it location/1}, {\it
  move/3}, {\it on/3}; a location is a {\it block} or the {\it
  table}. The constant {\it  maxsteps} is an upper bound on the length of a
plan. States of the domain are modeled by the ground atoms of the
form {\it on(b,l,t)} stating that block $b$ is at location $l$ at time $t$.
Actions are modeled by ground atoms {\it move(b,l,t)} stating that block $b$
is moved to location~$l$ at time $t$.

The {\gen} section of a program consists of a single rule
$$
\{\mi{move}(B,L,T)\}  \ar \mi{block}(B),\ \mi{location}(L),\ \mytime(T),\ T<\maxsteps.
$$
that defines a potential solution to be an arbitrary set of {\it move}
actions executed before {\it maxsteps}.

The fact that moving a block to a position at time $T$
forces a block  to be at this position at time $T{+}1$ is encoded in
{\define} part of the program by the rule
$$
\ba{l}
\mi{on}(B,L,T{+}1)
  \:{\ar}\:
  \mi{move}(B,L,T),\:\mi{block}(B),\:\mi{location}(L),\:\mytime(T),\:T{<}\maxsteps.\\
\ea
$$
The rule below specifies the commonsense law of inertia for a predicate {\it on}
stating that unless we know that the block is no
longer at the same position it remains where it was:
$$
\ba{ll}
\mi{on}(B,L,T{+}1) \ar &\mi{on}(B,L,T),\ \mi{not}\ \neg \mi{on}(B,L,T{+}1),\ \mi{block}(B),\
\mi{location}(L),\\
& \mytime(T),\ T<\maxsteps.\\
\ea
$$

The following constraint in {\test} encodes the restriction that
a block cannot be moved unless it is clear
$$
\ba{ll}
\ar & \mi{move}(B,\mi{L},T),\ \mi{on}(\mi{B1},B,T),\ \mi{block}(B),\ \mi{block}(\mi{B1}), \\
    & \mi{location}(L),\ \mytime(T),\ T<\maxsteps.
\ea
$$

Given the rest of the encoding and the description of an initial state
and of the goal state,
answer sets of the resulting program represent plans.
The ground atoms of the form {\it move(b,l,t)} present in an answer set form the list of
actions of a corresponding plan.

\subsection{Combinatory Categorial Grammar}\label{sec:revccg}

 Combinatory Categorial Grammar (CCG)~\cite{ste00}
is a linguistic grammar formalism.
Compared to other grammar formalisms,
CCG uses a comparatively small set of combinatory rules
-- combinators -- to combine comparatively rich lexical categories of words.

Categories in CCG are either atomic or
 complex. For instance,   noun $N$, noun phrase $\NP$,
 propositional phrase $\mi{PP}$, and sentence $S$ are atomic categories. Complex categories
 are functors that specify the type and direction of the
 arguments and the type of the result. A complex
 category
$$
S\bs \NP
$$
is a category for English intransitive verbs (such as {\sl walk}, {\sl hug}),
which states that a noun phrase is required to the left, resulting in
a sentence. A category
$$
(S\bs \NP)/\NP
$$
for English transitive verbs (such as {\sl like} and {\sl bite}) specifies
that a noun phrase is required to the right and yields the category of an
English intransitive verb, which (as before) requires a noun phrase to the left to form a sentence.

Given a sentence and a lexicon containing a set of word-category
pairs, we can replace words in the sentence by appropriate
categories. For example, for a sentence
\beq
\hbox{{\sl The dog bit John}}
\eeq{eq:s1}
and a lexicon containing pairs
\beq
\mi{The} \hbox{~-~} \NP/N;\  dog \hbox{~-~} N;\ bit \hbox{~-~} (S\bs\NP)/\NP;\ John \hbox{~-~} \NP
\eeq{eq:lexicon}
we obtain
\[
\ba{c@{\hspace{1cm}}c@{\hspace{1cm}}c@{\hspace{1cm}}c}
\infer{\NP/N}{The} &~~~  \infer{N}{dog} &~~~ \infer {(S\bs \NP)/\NP} {bit} &~~~ \infer{\NP}{John}
\hbox{.}
\ea
\]
Words may have multiple categories, e.g., ``bit'' is also an intransitive verb and a noun.
For presentation of parsing we limit each word to one category.
Our framework is able to handle multiple categories
by considering all combinations of word categories.

To parse English sentences a number of combinators are required~\cite{ste00}:
forward and backward application ($\fapp$ and $\bapp$, respectively),
forward and backward
{\sl composition} ($\fapp\Bc$ and $\bapp\Bc$),
forward and backward {\sl type raising}  ($\fapp\Tcc$ and $\bapp\Tcc$),
backward {\sl cross composition}, %
backward {\sl cross substitution}, %
and {\sl coordination}. %
Specifications of some of these combinators follow:
$$
\ba{c@{\hspace{1cm}}c@{\hspace{1cm}}c}%
\infer[\fapp] {A} {A/B \ \ \ B}&
\infer[\fapp\Bc] {A/C} {A/B \ \ \ B/C} &
\infer[\fapp\Tcc] {B/(B\bs A)} {A}\\
\infer[\bapp] {A} {B \ \ \ A\bs B} & \infer[\bapp\Bc] {A\bs C} {B\bs C\ \ \  A\bs B} &
\infer[\bapp\Tcc] {B\bs(B/A)} {A} %
\ea
$$
where $A$, $B$, $C$ are variables that can be substituted by CCG
categories such as $N$ or $S\bs \NP$.  An {\sl instance} of a CCG combinator
is obtained by  substituting CCG categories for variables.
For example,
\beq
\infer[\fapp] {\NP} {\NP/N \ \ \ N}
\eeq{ex:c1}
is an instance of the forward application combinator ($\fapp$).

A CCG combinatory rule combines one or more adjacent categories %
and yields exactly one output category. To parse a sentence is
 to apply instances of CCG combinators so that  the final
category $S$ is derived at the end.
A sample CCG derivation
for sentence~\eqref{eq:s1}
follows
\beq
\infer[\bapp] {S} {\infer[\fapp] {\NP}{ \infer{\NP/N}{The} &  \infer{N}{dog}}
              &
              \infer[\fapp]{S\bs \NP}{\infer {(S\bs \NP)/\NP} {bit} & \infer{\NP}{John}
}}\hbox{.}
\eeq{ex:dbj1}
Section~\ref{sec:parsingformal} gives a formal definition of the CCG parsing task.

\subsubsection{Type Raising and Spurious Parses:}

CCG restricted to
application combinators  generates the same language as
CCG restricted to application, composition, and type raising rules~%
\cite{1983_l_generalized_conjunction_and_type_ambiguity,1988_type_raising_functional_composition_and_non_constituent_conjunction}.
One of the motivations for type raising are non-constituent coordination constructions%
\footnote{E.g, in the sentence ``We gave Jan a record and Jo a book'',
neither ``Jan a record'' nor ``Jo a book'' is a linguistic constituent of the sentence.
With raising we can produce meaningful categories for these non-constituents
and subsequently coordinate them using ``and''.}
that can only be parsed with the use of raising~%
\cite[Example (2)]{2004_l_coordinate_ellipsis_and_apparent_non_constituent_coordination}.

Unrestricted applications of composition and type raising combinators often
create spurious parse trees which are semantically equivalent to
 parse trees derived using application rules only. Eisner~\cite[Example~(3)]{1996_l_efficient_normal_form_parsing_for_combinatory_categorial_grammar}
presents a sample sentence with 12 words and~252 parses but only 2 distinct meanings.
An example of a spurious parse for sentence~\eqref{eq:s1} is the following derivation
\beq
\infer[\fapp] {S}
          {\infer[\fapp\Bc]{S/\NP}{
                           \infer[\fapp\Tcc]{S/(S\bs \NP)}{
                                                   \infer[\fapp]{\NP}{\infer{\NP/N}{\mi{The}}
                                                                  &
													             \infer{N}{\mi{dog}}
												                }
						                           }
                           &
                           \infer{(S\bs \NP)/\NP}{\mi{bit}}
                           }
           &
			\infer{\NP}{\mi{John}}}
\eeq{eq:s1spurious}
which utilizes application, type raising, and composition combinators.
Both derivations~\eqref{ex:dbj1} and~\eqref{eq:s1spurious} have the
same semantic value (in a sense, the difference between~\eqref{ex:dbj1}
and~\eqref{eq:s1spurious} is not essential for subsequent semantic
analysis).%

In this work we aim at the generation of parse trees that
have different semantic values so that they reflect a real ambiguity of natural language,
and not a spurious ambiguity that arises from the underlying CCG formalism.
Various methods for dealing with spurious parses have been proposed
such as limiting type raising only to certain categories~\cite{cla03},
normalizing branching direction of consecutive composition rules
by means of predictive combinators~\cite{wit87}
or restrictions on parse tree shape~\cite{1996_l_efficient_normal_form_parsing_for_combinatory_categorial_grammar}.
We combine and extend these ideas
to pose restrictions on generated parse trees within our framework.
Details about normalizations and type raising limits that we implement %
are discussed in Section~\ref{sec:normalizations}.

\section{CCG Parsing via Planning}\label{sec:ccgviaplanning}
\subsection{Problem Statement}\label{sec:parsingformal}
We start by defining precisely the task of {\sl CCG parsing}. We then
state how this task can be seen as a planning problem.

A {\sl sentence} is  a sequence of words.
An {\sl abstract sentence representation} (ASR) is a sequence
of categories annotated by a unique \id.
Recall that
given a  lexicon,  we can replace words in the sentence by appropriate
categories. As a result we can turn any sentence into ASR using a
lexicon.
For instance, for  sentence~(\ref{eq:s1}) and lexicon~\eqref{eq:lexicon} a sequence
\beq
[\NP/N^1,~~ N^2,~~ (S\bs \NP)/\NP^3,~~ \NP^4].
\eeq{ex:asr1}
is an ASR of~(\ref{eq:s1}).  We refer to categories annotated by \id{}'s
as {\sl annotated categories}. Members of
\eqref{ex:asr1} are annotated categories.

Recall that an instance of a CCG combinator $C$ has a general form
$$
\infer[C] {Y}{X_1,\dots,X_n}\hbox{.}
$$
We say that the sequence $[X_1,\dots,X_n]$
is a {\sl precondition} sequence of $C$,
whereas $Y$ is an {\sl effect} of applying $C$.
The precondition sequence and the effect
of   instance \eqref{ex:c1} of the
combinator~$\fapp$
are $[\NP/N, N]$ and $\NP$,
respectively. Given an instance $C$ of a CCG combinator we may
annotate it by (i)  assigning a distinct {\id} to each member of its precondition
sequence, and (ii) assigning the $\id$ of the left most annotated
category in the precondition sequence to its effect. We say that such an
instance is an {\sl annotated (combinator) instance}.
For example,
\beq
\infer[\fapp] {\NP^1} {\NP/N^1 \ \ \ N^2}
\eeq{ex:ac2}
is an annotated instance w.r.t.  \eqref{ex:c1}.

We say that an annotated instance $C$ of a CCG combinator is {\sl relevant} to
an ASR sequence $A$ if the precondition sequence of $C$ is a substring
of $A$. An annotated instance~$C$ is applied to an ASR sequence $A$ by
replacing the substring of $A$ corresponding to the  precondition
sequence of $C$ by its effect. For example, annotated instance
\eqref{ex:ac2}  is relevant to ASR \eqref{ex:asr1}. Applying
\eqref{ex:ac2} to \eqref{ex:asr1} yields ASR
$[\NP^1,(S\bs \NP)/\NP^3,\NP^4]$.
In the following we will often say annotated combinator in place of
annotated instance.

To view CCG parsing as a planning problem we need to specify
 states and actions of this domain. In CCG planning,
states are ASRs and actions are annotated combinators. So the task is
given the initial ASR, e.g., $[X^1_1,\dots,X^n_n]$, to find a
sequence of annotated combinators that leads to the goal ASR --- $[S^1]$.

Let $\mathcal{C}_1$ denote annotated combinator \eqref{ex:ac2},
$\mathcal{C}_2$ denote
$$
\infer[\fapp] {S\bs\NP^3} {(S\bs\NP)/NP^3~~~\NP^4},
$$
and $\mathcal{C}_3$ denote
$$
\infer[\fapp] {S^1} {\NP^1 & S\bs \NP^3}.
$$
Given ASR \eqref{ex:asr1} a sequence of actions
$\mathcal{C}_1$, $\mathcal{C}_2$, and $\mathcal{C}_3$ forms a plan:
\beq
\ba{lll}
\hbox{Time 0:} && [\NP/N^1,~~ N^2,~~ (S\bs \NP)/\NP^3,~~ \NP^4] \\
~~~~~~\hbox {action:} & \mathcal{C}_1 &\\
\hbox{Time 1:} && [\NP^1,~~ (S\bs \NP)/\NP^3,~~ \NP^4],\\
~~~~~~\hbox {action:} & \mathcal{C}_2 &\\
\hbox{Time 2:} && [\NP^1,~~ S\bs \NP^3],\\
~~~~~~\hbox {action:} & \mathcal{C}_3 &\\
\hbox{Time 3:} && [S^1].
\ea
\eeq{eq:plan1}
This plan corresponds to  parse tree~\eqref{ex:dbj1} for sentence~\eqref{eq:s1}. On the other
hand, a plan formed by a sequence of actions $\mathcal{C}_2$,
$\mathcal{C}_1$, and $\mathcal{C}_3$ also corresponds to~\eqref{ex:dbj1}.

In planning the notion of {\sl serializability} is important.
Often given a plan, applying several consecutive actions in the
plan in any order or in parallel does not change the effect of their application.
Such plans are called {\sl serializable}.
Consequently, by allowing parallel execution of
actions one may represent a class of plans by a single one. This is a
well-known optimization in planning. For example, plan
$$
\ba{lll}
\hbox{Time 0:} && [\NP/N^1,~~ N^2,~~ (S\bs \NP)/\NP^3,~~ \NP^4] \\
~~~~~~\hbox {actions:} & \mathcal{C}_1, \mathcal{C}_2  &\\
\hbox{Time 1:} && [\NP^1,~~ S\bs \NP^3],\\
~~~~~~\hbox {action:} & \mathcal{C}_3 &\\
\hbox{Time 2:} && [S^1]
\ea
$$
may be seen as an abbreviation for a group of plans, i.e., itself,
plan~\eqref{eq:plan1}, and a plan formed by a sequence $\mathcal{C}_2$,
$\mathcal{C}_1$, and $\mathcal{C}_3$.
In CCG parsing as a planning problem, we are interested in finding
plans of this kind, i.e., plans with concurrent actions.

We note that the planning problem that we solve is somewhat different from the one we just
described as we would like to eliminate (``ban'') some of the plans
corresponding to spurious parses by enforcing normalizations.

\subsection{ASP Encoding}
In an ASP approach to CCG parsing, the goal is to encode
 the planning problem presented above as a logic
 program so that its answer sets correspond to plans. As a result
answer sets of this program will contain
the sequence of annotated combinators (actions, possibly concurrent) such that
the application of this sequence leads from a given ASR to the ASR
composed of a single category $S$.
We present a part of the encoding {\ccgasp}\footnote{The complete
 listing of {\ccgasp} is available at \\ \url{http://www.kr.tuwien.ac.at/staff/ps/aspccgtk/ccg.asp}}
 in the {\gringo} language that solves
a CCG parsing problem  by means of ideas presented in
Section~\ref{sec:asp}.

First, we need to decide how we represent states --- ASRs --- by sets
of ground atoms. To this end, we introduce symbols called ``positions''
that encode annotations of ASR members.
 In {\ccgasp}, relation $\posCat(p,c,t)$ states that a  category~$c$
 is annotated with (position) $p$ at time $t$.
Relation $\posAdjacent(p_L,$ $p_R,t)$ states that a position~$p_L$
is adjacent to a position  $p_R$ at time~$t$.
In other words, a category annotated by~$p_L$ immediately precedes
 a category annotated by $p_R$ in an ASR that
corresponds to a state at time~$t$ (intuitively, $L$
and $R$ denote left and right, respectively.)
These relations allow us to encode states of a CCG planning
domain.
For example, given an  ASR~\eqref{ex:asr1}
as the initial state,
we can encode this state by the following set of facts
\beq
\ba{l}
\posCat(1,\rfunc(``\NP",``N"),0).\ \ \posCat(2,``N",0).\\
\posCat(3,\rfunc(\lfunc(``S",``\NP"),``\NP"),0).\ \ \posCat(4,``\NP",0). \\
\posAdjacent(1,2,0).\  \ \posAdjacent(2,3,0).\  \ \posAdjacent(3,4,0).\
\ea
\eeq{eq:initstate}

Next we need to choose how we  encode  actions by ground atoms.
The combinators
mentioned in Section~\ref{sec:revccg} are of two kinds: the ones whose
precondition sequence consists of a single element (i.e., $\fapp\Tcc$
and $\bapp\Tcc$) and of two elements (e.g., $\fapp$ and
$\bapp$)\footnote{In fact, coordination combinator is of the third
  type, i.e., its precondition sequence contains three elements.
  Presenting the details of its encoding is out of the scope of this
  paper.}.
We call these
combinators {\sl unary} and {\sl binary} respectively.
Reification of actions is a  technique used in planning that
allows us to talk about common properties of actions in a compact way.
To utilize this idea, we first introduce relations $\unary(a)$ and $\binary(a)$
 for every  unary and binary combinator $a$
respectively.  For a unary combinator $a$, a relation $\occurs(a,p,c,t)$
states that a type raising action $a$  occurring at
time~$t$ raises a category identified with position $p$ (at time $t$)
to category $c$.
For a binary combinator $a$
a relation $\occurs(a,p_L,p_R,t)$
states that an action $a$ applied to positions~$p_L$ and $p_R$
occurs at time~$t$. For instance,  given the initial state~\eqref{eq:initstate}
\begin{itemize}
\item  $\occurs(\mi{ruleFwdTypeR},4,(S\bs \NP)/\NP,0)$
represents an application of the annotated combinator
$$
\infer[\fapp\Tcc]{(S\bs \NP)/\NP^4}{\NP^4}
$$
to~\eqref{eq:initstate} at time $0$,
\item $\occurs(\ruleFwdAppl,1,2,0)$ represents an application
  of~\eqref{ex:ac2} to~\eqref{eq:initstate} at time $0$.
\end{itemize}

Given an atom $\occurs(A,P,X,T)$ we sometimes say that an action $A$
{\sl modifies} a position $P$ at time $T$.

The {\gen} section of {\ccgasp} contains  the
rules of the kind
$$
\ba {ll}
\{ \occurs(\ruleFwdAppl, L,R,T) \} \ar
&	\posCat(L,\rfunc(A,B),T),\ \posCat(R,B,T),\ \\
& 	\posAdjacent(L,R,T),\\
&	\mi{not}\ \ban(\ruleFwdAppl, L,T),\\
& \mytime(T),\ T<\maxsteps.
\ea
$$
for each combinator.
Such choice rules describe a potential solution to the planning
problem as an arbitrary set of actions executed before {\it maxsteps}.
These rules also captures some of the executability conditions of
the corresponding actions.  For example,
$\posCat(L,\rfunc(A,B),T)$ states that the left member of the precondition
sequence of the forward application combinator  $\ruleFwdAppl$ is of
the form $A/B$. At the same time, $\posAdjacent(L,R,T)$ states that
$\ruleFwdAppl$ may be applied only to adjacent positions.
 A  relation $\ban(a,p,t)$
specifies when it is impossible for an action $a$ to modify
position~$p$ at time $t$.
Often there are several rules defining
this relation for a combinator. These rules form the main
mechanism by which normalization techniques are encoded in \ccgasp.
For instance, a rule defining $\ban$ relation
$$
\ba{ll}
\ban(\ruleFwdAppl,L,T)\ar&
 	\occurs(\ruleBwdRaise,L,X,\TLast{-}1),\\
& \posLastAffected(L,\TLast,T),\ \mytime(\TLast), \\
&	\mytime(T),\ T<\maxsteps.
\ea
$$
states that a forward application modifying a position $L$
 may not {\it occur} at time $T$  if the last action modifying $L$ was backward type
raising ($\posLastAffected$ is an auxiliary predicate that helps to
identify the last action modifying an element of the ASR).
This corresponds to one of the normalization rules discussed
in~\cite{1996_l_efficient_normal_form_parsing_for_combinatory_categorial_grammar}.

There are a number of rules that specify effects of actions in the CCG
parsing domain.
One such rule
$$
\ba{ll}
\posCat(L,A,T{+}1) \ar & \occurs(\ruleFwdAppl,L,R,T),
\\
&\posCat(L,\rfunc(A,B),T),\ \mytime(T),\ T<\maxsteps.
\ea
$$
states that an application of a forward application combinator at
time $T$
causes a category annotated by $L$ to be $X$ at time $T{+}1$.

The following rule characterizes an effect of binary combinators
and defines the {\it posAffected} concept which is useful in
stating several normalization conditions described in Section~\ref{sec:normalizations}:
$$
\ba{ll}
\posAffected(L,T{+}1) \ar
&\occurs(Act,L,R,T),\  \binary(Act),\\
&\ \mytime(T),\ T<\maxsteps.
\ea
$$
Relation $\posAffected(L,T{+}1)$ holds if the element annotated by $L$ in the ASR
was modified by a combinator at time  $T$.
Note that this rule takes advantage of reification and provides means
for compact encoding of common effects of all binary actions.
Furthermore, $\posAffected$ is used to state the law of inertia for the
predicate $\posCat$
$$
\ba{ll}
\posCat(P,C,T{+}1) \ar
&	 \posCat(P,C,T),\
	\mi{not}\ \posAffected(P,T{+}1),\\
&\mytime(T),\ T<\maxsteps.
\ea
$$

In the {\test} section of the program we encode such restrictions as no
two combinators may modify the same position simultaneously and
the fact that the goal has to be reached. We allow two possibilities for
specifying a goal. In one case, the goal is to reach an ASR composed of
a single category $S$ by $\maxsteps$. In another case, the goal is to
reach the shortest possible ASR sequence  by $\maxsteps$.

  Finally we pose additional restrictions, which ensure
that only a single plan is produced when multiple serializable plans
correspond to the same parse tree.
Note that applying a CCG rule $r$ at a  time $t$
  creates a new category required for subsequent application of another
  rule $r'$ at a time $t' {>} t$. We request that $r'$ is applied
  at $t'{=}t{+}1$.
Furthermore,  in \ccgasp we enforce the condition that
 combinators are applied as early as possible: %
by requesting that a rule applied at time $t$
  uses at least one position that was modified at time $t{-}1$.

Given \ccgasp and  the set of facts describing the initial state
(ASR representation of a sentence)
and  the goal state (ASR containing a single category $S$),
answer sets of the resulting program encode plans corresponding to
parse trees.
The ground atoms of the form $\occurs(a,p,c,t)$
present in an answer set form the list of
actions of a matching plan.

\subsection{Normalizations}\label{sec:normalizations}

Currently, \ccgasp implements a number of normalization techniques and
strategies for improving  efficiency and eliminating spurious parses:
\begin{myitemize}
\item
  One of the techniques used in \candc to improve its
  efficiency is to limit  type raising to certain categories {\sl
  based on the most commonly used type raising rule instantiations in
  sections 2-21 of CCGbank}~\cite{cla03}.
  We adopt this idea by limiting type raising to be applicable only to
  noun phrases, $\NP$,
	so that $\NP$ can be raised using categories $S$,
	$S\bs \NP$, or $(S\bs \NP)/\NP$.
  This technique reduces the size of the propositional (ground)
  program for \ccgasp and subsequently the performance of \ccgasp  considerably.
	We plan to extend  limiting type raising to the full set of
  categories used in \candc that proved to be
 suitable for wide-coverage parsing.

\item
  We normalize branching direction of subsequent functional composition operations~%
	\cite{1996_l_efficient_normal_form_parsing_for_combinatory_categorial_grammar}.
  This is realized by disallowing functional forward composition
  to apply to a category on the left side,
  which has been created by functional forward composition.
  (And similar for backward composition.)
\item
  We disallow certain combinations of rule applications if the same result
  can be achieved by other rule applications as shown in the following
  \medskip

  ~\;%
  \mbox{%
  \deriv{3}{
  X/Y & Y/Z & ~Z \\
  \fcomp{2} & \\
  \mcc{2}{X/Z} \\
  \fapply{3} \\
  \mcc{3}{X}
  }%
  \;
  \raisebox{-3em}{$\stackrel{%
    {\rotatebox{90}{\scriptsize{}\ \raisebox{0.5em}{normalize}}}}{%
    {\Rightarrow}}$}
  \;
  \deriv{3}{%
  X/Y & ~Y/Z & Z \\
   & \fapply{2} \\
   & \mcc{2}{Y} \\
  \fapply{2} \\
  \mcc{2}{X}
  }
  }
  \quad
  \mbox{
  \deriv{2}{
  X & ~Y\bs X \\
  \ftype{1} & \\
  Y/(Y\bs X) \\
  \fapply{2} \\
  \mcc{2}{Y}
  }%
  \;
  \raisebox{-3em}{$\stackrel{%
    {\rotatebox{90}{\scriptsize{}\ \raisebox{0.5em}{normalize}}}}{%
    {\Rightarrow}}$}
  \;
  \deriv{2}{
  X & ~~Y\bs X \\
  \bapply{2} \\
  \mcc{2}{Y}
  }$\!\!$%
  }%

  \smallskip
  where the left-hand side is the spurious parse and the right-hand side the normalized parse.
  These two normalizations
  (plus analogous normalizations for backward composition and backward type raising)
  eliminate spurious parses like~\eqref{eq:s1spurious}
  and have been discussed in similar form in~%
  \cite{2004_l_type_inheritance_combinatory_categorial_grammar,1996_l_efficient_normal_form_parsing_for_combinatory_categorial_grammar}.
\end{myitemize}

\section{ASPCCG Toolkit}\label{sec:implframework}

\newcommand\blockdiagram{%
    \beginpgfgraphicnamed{blockdiagrampdf}
    \begin{tikzpicture}[trapezium left angle=80,trapezium right angle=100]%
    \node[rectangle,draw] (candc) at (0,0)
      {\candc supertagger
      };
    \node[rectangle,draw,minimum height=3.8em,
			below=2em of candc] (gcparse)
      {\gringo + \clasp};
    \node[rectangle,draw,below=2em of gcparse] (viz)
      {\begin{tabular}{@{}c@{}}
			 \gringo + \clasp \\
			 + \idpdraw
			 \end{tabular}};
    \draw[->]
			($(candc.south)-(1,0)$) -- ($(gcparse.north)-(1,0)$);
		\node[anchor=west] (candcout) at
    	($($(candc.south)-(1,0)$)!0.5!($(gcparse.north)-(1,0)$)$)
			{Sequence of words + category tags for each word};
    \draw[->]
			($(gcparse.south)-(1,0)$) -- ($(viz.north)-(1,0)$);
		\node[anchor=west] at
    	($($(gcparse.south)-(1,0)$)!0.5!($(viz.north)-(1,0)$)$)
			{Parser answer sets};
    \node[trapezium,draw,anchor=north east] (inputsent)
      at ($(candc.north west)+(-2.0,0)$)
      {Sentence (string)};
    \draw[->] (inputsent.east) -> (candc.west);
    \node[anchor=north,below=0.2em of inputsent] (or)
      {OR};
    \node[trapezium,draw,anchor=north,below=0.2em of or] (inputseq)
      {$\begin{array}{@{}c@{}}
			 \text{Sequence of words} \\
			 + \\
			 \text{Dictionary}
			 \end{array}\!\!$
			};
    \draw[->] (inputseq.east) ->
			($(gcparse.north west)!(inputseq.east)!(gcparse.south west)$);
    \node[trapezium,draw,anchor=east] (guiviz)
      at ($(viz.west)+(-1.0,0)$)
      {Visualization};
    \draw[->] (viz.west) -- (guiviz.east);
    \node[trapezium,draw,anchor=west] (parserenc)
      at ($(gcparse.east)+(0.6,0)$)
      {\ccgasp};
    \node[trapezium,draw,anchor=west] (vizenc)
      at ($(viz.east)+(0.6,0)$)
      {\ccgtoidpdrawasp};
    \draw[->] (parserenc.west) -- (gcparse.east);
    \draw[->] (vizenc.west) -- (viz.east);
		\coordinate(fwnw) at ($(candc.north west)+(-0.2,0.6)$);
		\node[rectangle,anchor=south,minimum width=1em]
			(vizoffset) at ($(viz.south)+(0,-0.2)$) {};
		\coordinate(fwsw) at
			($(vizoffset.south west)!(fwnw)!(vizoffset.south east)$);
		\coordinate(fwse) at
			($(vizoffset.south west)!($(candcout.east)+(0.25,0)$)!(vizoffset.south east)$);
		\coordinate(fwne) at
			($(fwnw)!(fwse)!($(fwnw)+(1,0)$)$);
		\draw[dashed,line width=1pt]
			(fwnw) -- (fwne) -- (fwse) -- (fwsw) -- cycle;
    \node[anchor=north] (pythonframework)
      at ($(fwnw)!0.5!(fwne)$)
      {\aspccg};
    \end{tikzpicture}%
    \endpgfgraphicnamed
	}

We have implemented \aspccg --- a python%
\footnote{{\tt http://www.python.org/}}
framework for using \ccgasp.
The framework is available online%
\footnote{{\tt http://www.kr.tuwien.ac.at/staff/ps/aspccgtk/}},
including  documentation and examples.

Figure~\ref{fig:blockdiag} shows a block diagram of \aspccg.
We use \gringo and \clasp for ASP solving
and control these solvers from python using a modified version
of the BioASP library~%
\cite{2010_the_bioasp_library_asp_solutions_for_systems_biology}.
BioASP is used for calling ASP solvers as subtasks,
parsing answer sets, and writing these answer sets to temporary files as facts.

Input for parsing can be
\begin{inparaenum}[(a)]
\item
	a natural language sentence given as a string, or
\item
	a sequence of words
	and a dictionary providing possible categories for each word,
	both given as ASP facts.
\end{inparaenum}
In the first case, the framework uses \candc supertagger%
\footnote{{\tt http://svn.ask.it.usyd.edu.au/trac/candc}}~%
\cite{2004_l_parsing_the_wsj_using_ccg_and_log_linear_models} to tokenize and tag this sentence.
The result of supertagging is a sequence of words of the sentence, where
 each word is assigned a set of likely CCG categories.
From the \candc supertagger output, \aspccg creates a set of ASP facts
representing the sequence of words and a
 corresponding set of likely CCG categories. %
This set of facts is passed to \ccgasp as the initial state.
In the second case (b) the input can be processed directly by \ccgasp.
The maximum parse tree depth (i.e., the maximum plan length -- {\it maxsteps})
currently has to be specified by the user.
Auto detection of useful depth values is subject of future work.

\aspccg first attempts to find a ``strict'' parse
which requires that the resulting parse tree
yields a category $S$ (by $\maxsteps$).
If this is not possible, we do ``best-effort'' parsing
using \clasp optimization features %
to minimize the number of categories left at the end.
For instance, consider a lexicon that provides a single category for
``bit'', namely $(S\bs \NP)/\NP$, then the following derivation
\beq
\text{
\deriv{3}{
\mi{The} & \mi{dog} & \mi{bit} \\
\uline{1} & \uline{1} & \uline{1} \\
\NP/N & N & (S\bs \NP)/\NP \\
\fapply{2} & \\
\NP & \\
\ftype{1} & \\
S/(S\bs \NP) & \\
\fcomp{3} \\
& S/\NP
}
}
\eeq{ex:thedogbit}
corresponds to a best-effort parse.

\begin{figure}[btph]%
	\begin{center}%
	\includegraphics[width=\linewidth]{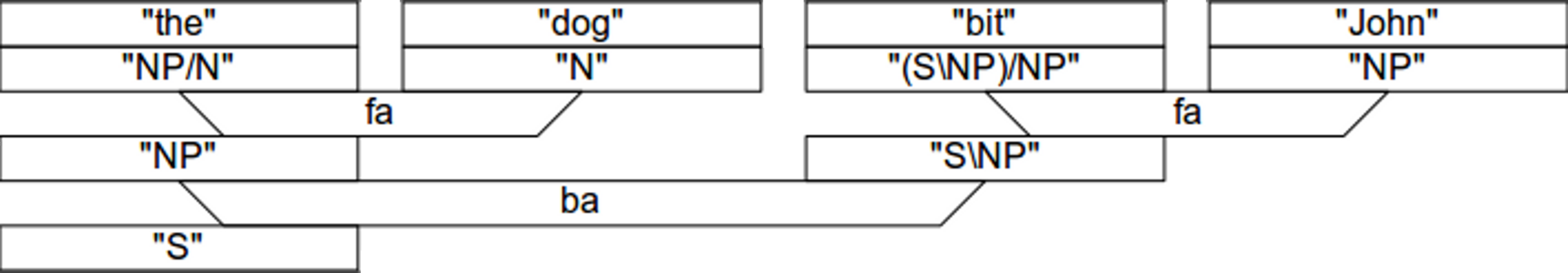}%
	\end{center}%
  \caption{Visualization of parse tree~\eqref{ex:dbj1} for sentence~\eqref{eq:s1} using \idpdraw.}%
  \label{fig:idpdrawoutput}%
\end{figure}

Answer sets resulting from \ccgasp
represent parse trees.
\aspccg  passes them to a visualization component,
which invokes {\gringo}+{\clasp} on another ASP encoding \ccgtoidpdrawasp.%
\footnote{%
This visualization component could be put directly into \ccgasp.
However, for performance reasons it has proved crucial
to separate the parsing calculation from the drawing calculations.%
}
The resulting answer sets of \ccgtoidpdrawasp contain drawing instructions
for the \idpdraw tool~\cite{2009_idpdraw_manual}, which is used to produce
 a two-dimensional image for each parse tree.
Figure~\ref{fig:idpdrawoutput} demonstrates an image generated by
\idpdraw for parse tree~\eqref{ex:dbj1}
of sentence~\eqref{eq:s1}.
If multiple parse trees exist,
\idpdraw allows to switch between them.
\begin{figure}[btph]%
	\begin{center}%
  \blockdiagram%
	\end{center}%
  \caption{Block diagram of the ASPCCG framework. (Arrows indicate data flow.)}%
  \label{fig:blockdiag}%
\end{figure}

\section{Discussion and Future Work}\label{sec:discussion}

Preliminary experiments on using the \candc supertagger as a front-end
of \aspccg
yielded promising results for achieving wide-coverage parsing.
The supertagger of \candc not only provides a set of likely category
assignments for the words in a given sentence but also includes
probability values for assigned categories.
\candc uses a dynamic tagging strategy for parsing.
First only very likely categories from the tagger are used for parsing.
If this yields no result then less likely categories are also taken into account.
In the future, we will implement a  similar approach in \aspccg.

We have evaluated the efficiency of \aspccg
on a small selection of examples
from \ccgbank~\cite{2007_l_ccgbank}.
In the future we will evaluate our parser against a larger corpus of \ccgbank,
considering both parsing efficiency and quality of results as evaluation criteria.
Experiments done so far are encouraging
and we are convinced that wide-coverage CCG parsing using ASP technology
is feasible. %

To increase parsing efficiency we consider to reformulate the CCG parsing problem
as a ``configuration'' problem.
This might improve performance.
At the same time the framework would keep its beneficial declarative
nature.
Investigating applicability of incremental ASP~\cite{geb08} to enhance system's
performance is another direction of future research.

Creating semantic representations for sentences is an important task in natural
language processing.
\boxer~\cite{2008_boxer} is a tool which accomplishes this task,
given a CCG parse tree from \candc.
To take advantage of this advanced computational semantics tool, we
aim at creating an output format for \aspccg that is compatible with \boxer.

As our framework is a generic parsing framework,
we can easily compare different CCG rule sets
with respect to their efficiency and normalization behavior.
We next discuss an  %
idea for improving scalability of \ccgasp
that is based on an alternative combinatory rule set to the one currently
implemented in \ccgasp.
Type raising is a core source of nondeterminism in CCG parsing
and is one of the main reasons for spurious parse trees and  long parsing times.
In the future we would like to evaluate an approach that partially eliminates type raising
by pushing it into all non-type-raising combinators.
A similar strategy has been proposed for composition combinators by
Wittenburg~\cite{wit87}.%
\footnote{Wittenburg introduced a new set of combinatory rules
by combining the functional composition combinators with other combinators.
By omitting the original functional composition combinators,
certain spurious parse trees can no longer be derived.}
Combining CCG rules this way creates  more combinators,
however these rules contain fewer nondeterministic guesses about raising categories.
The reduced nondeterminism should improve solving efficiency
without losing any CCG derivations.

\noindent
\textbf{Acknowledgments.}
We would like to thank John Beavers and  Vladimir Lifschitz for
valuable detailed comments on the first draft of this paper.
We are grateful  to Jason Baldridge,  Johan Bos, Esra Erdem, Michael
Fink, Michael
Gelfond, Joohyung Lee, and Miroslaw Truszczynski
for useful discussions related to the topic of this
work, as well as to the anonymous reviewers for their feedback.
Yuliya Lierler was supported by a CRA/NSF 2010 Computing
Innovation Fellowship.
Peter Sch\"{u}ller was supported by the
Vienna Science and Technology Fund (WWTF) project ICT08-020.

\bibliographystyle{splncs03}

\end{document}